\documentclass{article}
\usepackage{spconf,amsmath,graphicx}
\usepackage{hyperref}
\usepackage{times}
\usepackage{epsfig}
\usepackage{graphicx}
\usepackage{amsmath}
\usepackage{amssymb}
\usepackage{algorithm} 
\usepackage{algpseudocode}  

\usepackage[square,numbers]{natbib}
\usepackage{times,epsfig,graphicx,amsmath,amssymb}
\usepackage{times,graphicx,amsmath,amssymb,booktabs,tabulary,multirow,overpic,bbm,amssymb}

\newcolumntype{x}[1]{>{\centering\arraybackslash}p{#1pt}}

\newlength\savewidth

\makeatletter\renewcommand\paragraph{\@startsection{paragraph}{4}{\z@}
  {.5em \@plus1ex \@minus.2ex}{-.5em}{\normalfont\normalsize\bfseries}}\makeatother
  
\usepackage{grffile}
\usepackage{amsmath}

\DeclareMathOperator*{\argmin}{arg\,min}
\usepackage{kotex}
\usepackage{multirow}
\usepackage{array}
\makeatletter
\newcommand{\thickhline}{%
    \noalign {\ifnum 0=`}\fi \hrule height 1.1pt
    \futurelet \reserved@a \@xhline
}
\newcolumntype{"}{@{\hskip\tabcolsep\vrule width 1pt\hskip\tabcolsep}}
\makeatother
\usepackage{subcaption}
\begin{document}

\pagestyle{plain}

\title{Blur Invariant Kernel-Adaptive Network \\for Single Image Blind deblurring}

\name{Sungkwon An$^{1}$ \thanks{$^{1}$These authors contributed equally to this study},
Hyungmin Roh$^{1}$ \footnotemark[1],
Myungjoo Kang$^{2}$\thanks{$^{2}$Corresponding author}}
\address{$^{1}$Computational Science and Technology, Seoul National University
\\$^{2}$ Department of Mathematics, Seoul National University
\\\textit{sk\_an@snu.ac.kr,  raingold1347@snu.ac.kr,  mkang@snu.ac.kr}
}

\maketitle
\thispagestyle{empty}

\begin{abstract}
 We present a novel, blind, single image deblurring method that utilizes information regarding blur kernels. Our model solves the deblurring problem by dividing it into two successive tasks: (1) blur kernel estimation and (2) sharp image restoration. We first introduce a kernel estimation network that produces adaptive blur kernels based on the analysis of the blurred image. The network learns the blur pattern of the input image and trains to generate the estimation of image-specific blur kernels. Subsequently, we propose a deblurring network that restores sharp images using the estimated blur kernel. To use the kernel efficiently, we propose a kernel-adaptive AE block that encodes features from both blurred images and blur kernels into a low dimensional space and then decodes them simultaneously to obtain an appropriately synthesized feature representation. We evaluate our model on REDS, GOPRO and Flickr2K datasets using various Gaussian blur kernels. Experiments show that our model can achieve state-of-the-art results on each dataset.
\end{abstract}

\section{Introduction}

\begin{figure*}[t]
	\begin{center}
		\includegraphics[width=\textwidth]{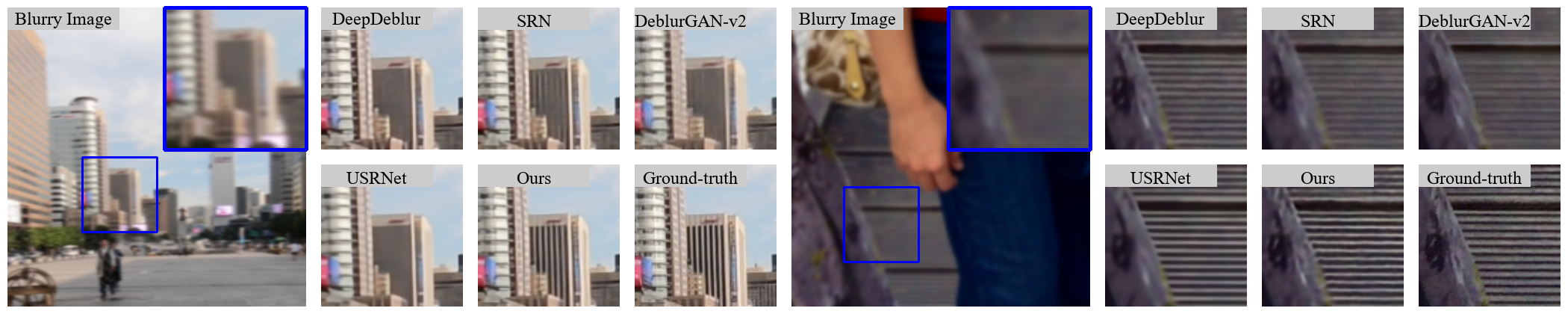}
		\caption{Visual example of image deblurring results. Most methods failed to restore complex patterns accurately. Our model successfully restores detailed patterns and the result appears similar to the ground-truth.}
		\label{fig:front_image}
	\end{center}
\end{figure*}

Many studies have been performed to solve the single image deblurring problem, and most previous approaches can be categorized into two groups: non-blind and blind methods. The non-blind method utilizes information from a blur kernel, whereas the blind method restores a sharp image without using any information from a blur kernel.

Although recent studies based on both methods have demonstrated good results, the two methods still present some limitations. First, when applying the non-blind method, known blur kernels are assumed. However, blur kernels are not known in most low-quality images; therefore, it is difficult to apply the non-blind method to real-world problems. Meanwhile, because the blind method does not require information regarding the blur kernel, it can be expanded limitlessly to a real-world problem.

Most state-of-the-art single image deblurring methods have adopted the blind method, which uses only a blur image to apply to the real-world. The blind method was developed owing to the difficulty in obtaining information regarding a blur kernel from arbitrary images. Nah \textit{\textit{et al.}}~\cite{Nah_2017_CVPR} proposed a deblurring network called DeepDeblur, which does not require knowledge regarding blur kernels. They opted not to include that information because incorrectly predicted kernels may introduce unintended artifacts to the results. However, if kernel information is accurately predicted, the performance can be further enhanced.
From that point of view, studies~\cite{kernelfusion, gong2017self, pan2019phase, networkpair} predicting a blur kernel and using it for image restoration has been proposed.


In this work, we propose a Blur Invariant Kernel-Adaptive Network (BIKAnet) to address the limits of previous methods and further improve the results. Our model comprises two stages. The first is a kernel estimation network that predicts the blur kernel of a blurred image. The second is a deblurring network that restores sharp images using information regarding the estimated kernel. To reconstruct images accurately, we first estimate suitable blur kernels by analyzing the blurred images. Subsequently, our deblurring network restores sharp images by analyzing the detail and structure of the images using the information from predicted blur kernels. The architectural design of the deblurring network is based on non-blind methods, in that adaptive blur kernels are used. However, when combined with our kernel estimation network, our BIKAnet can be applied to arbitrary images as a blind method. Furthermore, we propose the kernel-adaptive AE block and long-term skip connection that enable blurred images to be analyzed without any loss in coarse features from original images when the network deepens.

We created datasets using various anisotropic Gaussian kernels to validate our model on datasets that are close to real-world images rather than using existing diversity-limited datasets. Our model demonstrated state-of-the-art performance on data constructed for our purposes. We present our method to predict the blur kernel in Section \ref{sec:propsed}, where we also introduce a state-of-the-art BIKAnet for single image deblurring problems.

Our contribution is as follows:
\begin{itemize}
    \item We introduce a novel BIKAnet that can analyze coarse features from blurred images in a deep network and utilize estimated blur kernels.
    
    \item We propose a kernel-adaptive AE block to improve the understanding of the image deblurring process and the method to use blur kernel estimation efficiently. 
    
    \item We propose a long-term skip connection that allows our network to consider all progressively analyzed features from fine to coarse features.
    
    \item By combining the kernel estimation and BIKAnet, we introduce an end-to-end deblurring network pipeline that is not only applicable to real-world problems (like blind methods), but also yield accurate results (like non-blind methods).
\end{itemize}

\section{Related Studies}
In this section, we provide a brief review of some non-blind and blind approaches for solving the image deblurring problem.
\begin{figure*}[t]
    \begin{center}
    \includegraphics[width=\textwidth]{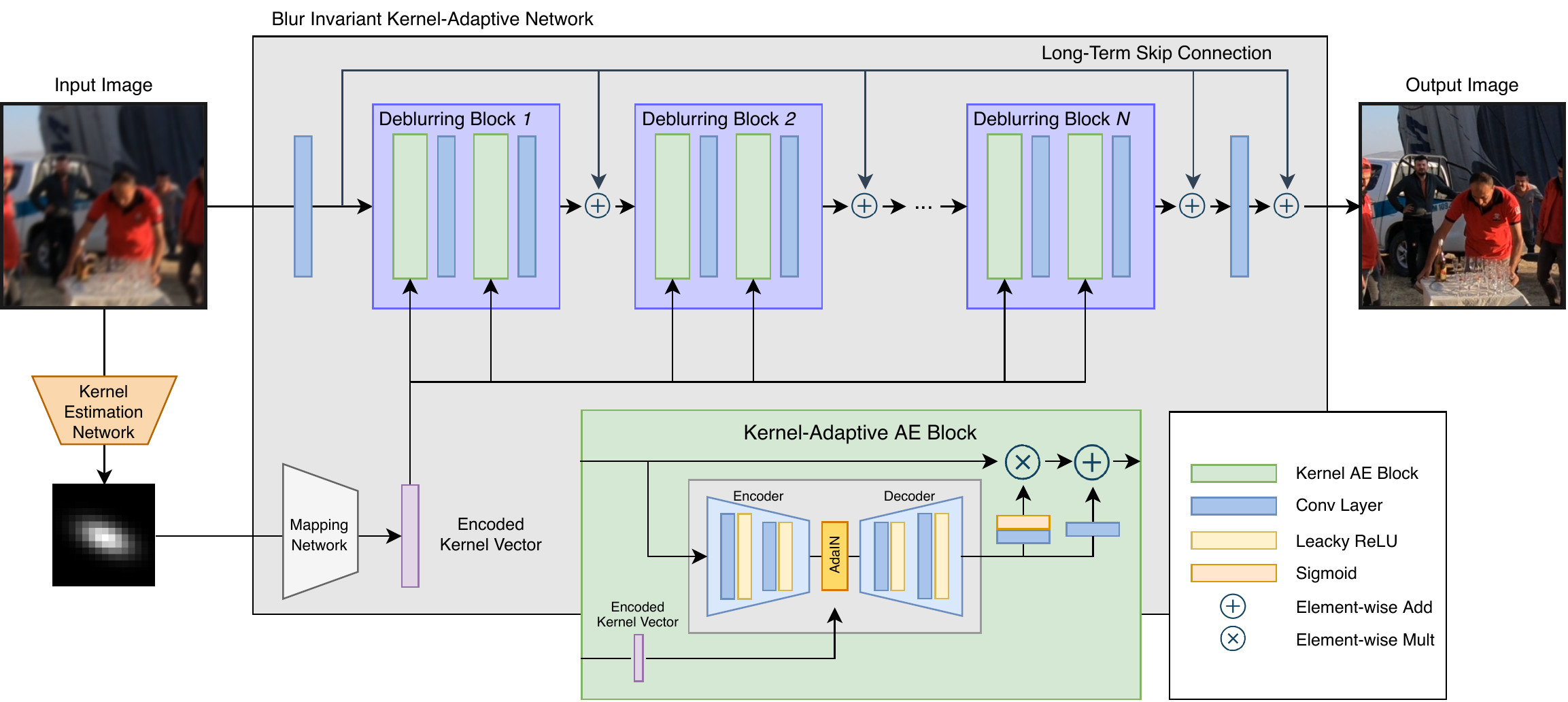}
    \caption{Our Network Architecture}
    \label{fig:network_pipeline}
    \end{center}
\end{figure*}

\subsection {Non-blind methods}

The non-blind method of restoring a sharp image using blur kernel information has been investigated continuously. 
Researchers have attempted to combine recent deep learning techniques and the traditional Wiener filter~\cite{wiener} to deconvolve images using blur kernels. Kruse \textit{et al.}~\cite{kruse2017learning} proposed an FFT-based non-blind deblurring method that utilizes an improved Wiener filter for a specified blur kernel and a CNN-based term as input. However, the circular blur assumption is required for an efficient FFT-based optimization. Wang \textit{et al.}~\cite{wang2018training} proposed a method to manage different types of blur kernels in a single model. They estimated the residuals from a CNN model after predeconvolving blur images using the regularized Wiener method.

Furthermore, Zhang \textit{et al.}~\cite{matconvnet} used a deep learning technique to perform iterative denoising and deblurring using the FCNN with a deconvolution module. Ren \textit{et al.}~\cite{lowrank} approximated an arbitrary pseudo-inverse kernel as multiple matrices and initialized the convolutional parameters of a deblurring network using these matrices. Their initialization method enabled the network to avoid poor local minima and yield a favorable deblurring performance. Vasu \textit{et al.}~\cite{handling} attempted to utilize both recent deep learning techniques and the hyper Laplacian prior to handle kernel uncertainty in non-blind motion deblurring. Lately, Nan and Ji~\cite{uncertainty} proposed model that handles kernel/noise error well and analyzed the impact of kernel/model error. Zhang \textit{et al.}~\cite{usrnet} took advantage of the information of the kernel by receiving the blur kernel as input and imposing the blurring constraint on the solution.

\subsection {Blind methods}

More recent works focus on the blind method, which is applicable to any blurred image even if no information exists regarding its blur kernel. Because the blind method does not require information regarding the blur kernel in the image restoration process, it can be extended to real-world problems more effectively than the non-blind method.

Because the application of multiscale architectures to various image processing tasks have demonstrated good performances, they have also been applied to the deblurring task. Nah \textit{et al.}~\cite{Nah_2017_CVPR} proposed a multiscale CNN called DeepDeblur, which restores a sharp image step by step by applying networks with the same structure from the coarsest level to a finer level and concatenating sharp features from the previous network. Similarly, Tao \textit{et al.}~\cite{srn} restored sharp images at multiple resolutions through a coarse-to-fine scale analysis. In a different attempt, Kupyn \textit{et al.}~\cite{deblurgan} proposed DeblurGAN with a conditional GAN framework and content loss. Additionally, they proposed DeblurGAN-v2~\cite{deblurganv2} with an FPN applied to a generator and global and local discriminators. Shen \textit{et al.}~\cite{human} proposed a triple-branch encoder--decoder structure that sharpens the foreground and background and analyzed them globally.

For the blind image enhancement method, researchers have attempted to estimate the degradation kernel and put it into the non-blind method to restore the clean image. Pan \textit{et al.}~\cite{darkchannel} applied motivation to deblurring by obtaining sparse dark pixels in the blurring process. Blur kernels were estimated by calculating the dark channel using a linear method; subsequently, the estimated kernel and the existing deblurring method were used to obtain a restored image. In a slightly different manner, they~\cite{pcode} also proposed a method to estimate blur kernels by exploiting reliable edges from blurred images and removing outliers from intermediate latent images. Furthermore, a robust energy function was proposed to develop an advanced deblurring algorithm.


Recently, Ren \textit{et al.}~\cite{deepprior} proposed SelfDeblur method that consists of two generative networks which capture the deep priors of blur kernel and sharp image, respectively. They used the self-supervised learning to optimize their networks using only a test image and no ground-truth sharp image. Kaufman and Fattal~\cite{networkpair} predicted the blur kernel by using the cross-correlation of the blur image convolution feature, and showed that using it in the restoration process brings performance improvement.



\section{Proposed Method}
\label{sec:propsed}
In this section, we provide the problem formulation and the architectural design of our proposed networks for blur kernel estimation and the image deblurring problem.

\subsection {Problem Formulation}

In the basic image deblurring problem, a sharp image $S$ and a blurred image $B$ are expressed as follows:
\begin{equation} \label{eq:0}
    B = k*S + n,
\end{equation}
where $k$ is the blur kernel, $n$ the additive noise, and $*$ the convolution operator. Kernel $k$ provides various blur effects, such as motion blur and Gaussian blur depending on its type.
In this problem, most methods estimate a deblurred image using a network that renders a blurred image sharp. We extend the deblurring problem to the problem of predicting kernel $k$, which is the cause of blurring, and use it to obtain a sharp image $S$. Obtaining the kernel that causes a blur image in the deblurring problem and using that information will not only yield more advanced results compared with existing methods, but will also enable us to understand the deblurring process. Equations~\eqref{eq:1} and ~\eqref{eq:2} represent the kernel estimation and deblurring mechanism, respectively.

\begin{align}
    \theta^*_g &= \argmin_{\theta_g}||\mathcal{G}\left(B, \theta_g\right) * S - B|| \label{eq:1}\\
    B &\approx \mathcal{G}\left(B, \theta_g^*\right) * S = k^* * S,
\end{align} where $\mathcal{G}$ is the kernel estimation network, $\theta_g$ denotes the parameters of $\mathcal{G}$, and $k^*$ is the estimated blur kernel.

\begin{align}
    \theta^*_n &= \argmin_{\theta_n}||\mathcal{N}\left(B, k^*, \theta_n\right) - S||\\
    &=\argmin_{\theta_n}||\mathcal{N}\left(B, \mathcal{G}(B, \theta^*_g), \theta_n\right) - S|| \label{eq:2},
\end{align} where $\mathcal{N}$ is the deblurring network, and $\theta_n$ denotes the parameters of $\mathcal{N}$.

We focused on deblurring images blurred by various Gaussian kernels. Additional experimental methods and results of motion blur kernels obtained using our method are provided in Section \ref{sec:experiments}.

\subsection {Blur Kernel Estimation Network}

\begin{figure}
    \begin{center}
    \includegraphics[width=.474\textwidth]{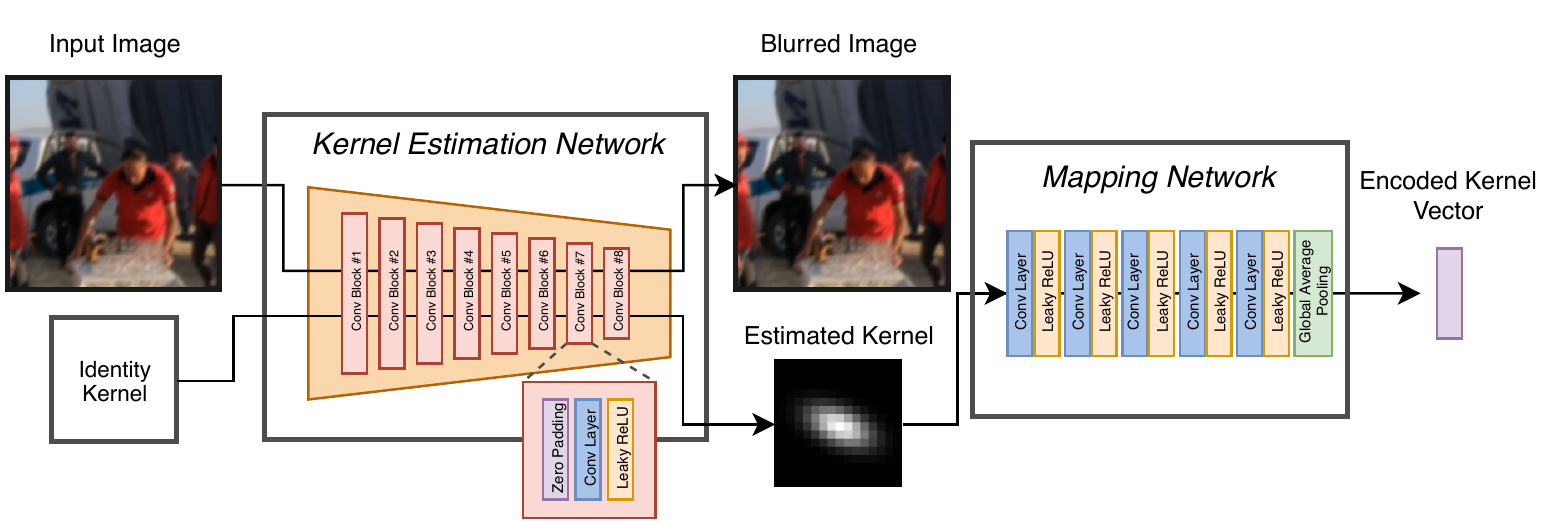}
    \caption{Kernel Estimation Network Architecture}
    \label{fig:kernel_estimator}
    \end{center}
\end{figure}

By estimating the blur kernel during the deblurring process, we can understand the blur information and obtain more accurate deblurred results. KernelGAN~\cite{kernelgan} estimates the degradation kernel that generates a low-resolution image by utilizing GAN's ability to learn data representation. We adopted KernelGAN's pipeline to estimate the blur kernel of a blurred image. The architecture is shown in Figure~\ref{fig:kernel_estimator}. KernelGAN estimates the downscaling kernel by learning the internal distribution of patches from low-resolution images. Its network contains a generator that downscales the input image and a discriminator that determines the distribution of patches in the input image.

We modified the KernelGAN structure by replacing the downscaling generator with our blurring generator such that the model fitted our task. We trained the discriminator to distinguish between the blur information of the input image and the fake blur image generated by the generator, whereas the blurring generator was trained to generate a fake blur image that can share the blur pattern of the input image such that it can deceive the discriminator. Consequently, we were able to extract the blur kernel of the input image using our blurring generator.

\subsection {Blur Invariant Kernel-Adaptive Network}

Figure~\ref{fig:network_pipeline} illustrates the pipeline of BIKAnet.
We opted not to reduce the size of the feature maps from the input images by avoiding the usage of pooling layers to maintain spatial information from blurred images. Hence, when applying the kernel feature to the blur image feature as a condition, an imbalance problem occurs for each feature size. Therefore, we introduced a kernel-adaptive autoencoder block, which comprises an encoder and a decoder, inspired by the conditional autoencoder.

We equip the Auto-encoder named kernel-adaptive AE block with Adaptive Instance Normalization (AdaIN)~\cite{adain} whose parameters are generated by a mapping network from the estimated kernel. The AdaIN parameters are dynamically generated for each kernel-adaptive AE block. Blocks constructed in this way use the kernel guided feature in the deblurring process.



When BIKAnet processes information, it passes through several kernel-adaptive AE blocks with a large number of layers. In this process, information extracted from the initial layers is likely to be lost; this is often observed in other neural networks with deep layers. To solve this problem, many studies utilize residual blocks with skip connections. Despite the usage of this connection, however, the consideration of coarse features in the image is limited if the number of blocks is too large and too deep. We avoided such information loss by passing the coarse level features to the middle of all deblurring blocks using a long-term skip connection. Through this, we solved the vanishing information problem and obtained better results. We checked the effect of the long-term skip connection on the performance improvement through ablation study in Section \ref{sec:experiments}.


 
\section{Experiments}

\label{sec:experiments}

In this section, we present the experimental condition and results by comparing the performances of our model and other models on three other dataset.

\subsection{Implementation Details}

We trained our model on REDS~\cite{Nah_2019_CVPR_Workshops_REDS}, GOPRO~\cite{Nah_2017_CVPR}, and Flickr2K~\cite{flickr} datasets. From the REDS dataset, we used 24,000 images for training and 3,000 for testing. For the GOPRO dataset, we used 2,103 images for training and 1,111 for testing. To train our model on the Flickr2K dataset, we randomly cropped HR images to increase the number of data and used 20,000 and 3,000 images for training and testing, respectively.

Our model was trained on Titan X and Titan RTX. We used the ADAM optimizer with a learning rate of $2\times 10^{-4}$ and trained for $200k$ iterations. 

\subsection {Data Generation}

\begin{figure}
    \begin{center}
    \includegraphics[width=.474\textwidth]{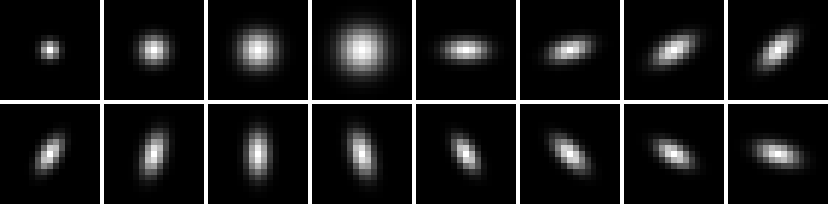}
    \caption{Sixteen blur kernels. We used 4 Gaussian kernels with various variances and 12 anisotropic Gaussian kernels with various rotation angles.}\label{fig:blur_kernels}
    \end{center}
\end{figure}

Most existing deblurring datasets tend to focus on the motion blur. Because we aim to train and test our model on a Gaussian blur, we created new datasets using our own blur kernels. 
Among the existing motion blur datasets, we used sharp images from REDS and GOPRO, and we used high-resolution images from the Flickr2K Super-Resolution dataset. Whereas REDS and GOPRO are composed of daily images from various scenes, Flickr2K contains detailed texture and delicate patterns. Therefore, these datasets are suitable for evaluating the performances of deblurring networks. 

We used $17 \times 17$ isotropic Gaussian kernels and anisotropic Gaussian kernels, both of which provide Gaussian and motion blur effects to sharp images. 
Each Gaussian blur kernel was created with different variance sizes and anisotropic Gaussian blurs of different rotations. 

Figure~\ref{fig:blur_kernels} illustrates the blur kernels we used in this study. Each kernel was randomly selected from 16 blur kernels and applied to each sharp image. Figure~\ref{fig:estimated_kernels} shows the blurred images generated by our blur kernels and the corresponding estimated kernels predicted by the kernel estimation network. Therefore, we were able to obtain data pairs comprising a blur image, a sharp image, a blur kernel, and an estimated kernel, which we used to train our model. 
\subsection {Kernel Estimation}

\begin{figure}
    \begin{center}
        \begin{subfigure}{.112\textwidth}
            \centering
            \includegraphics[width=\textwidth]{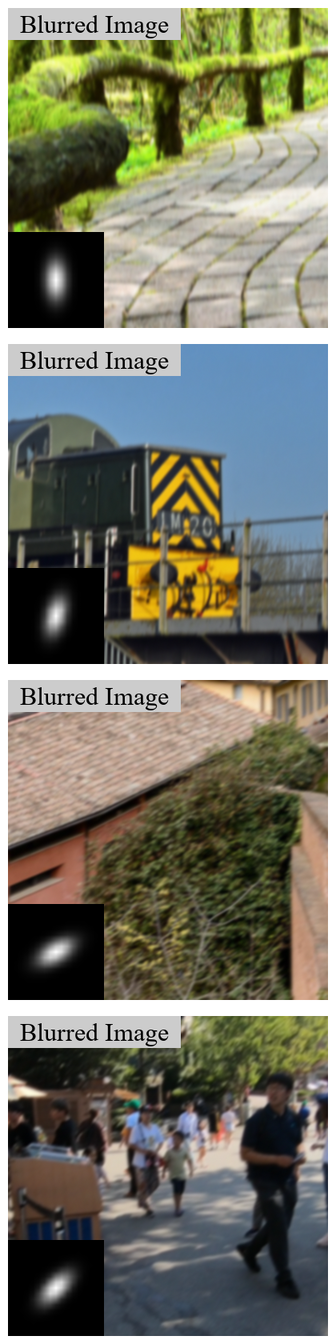}
            \caption{}
            \label{fig:deblur_a}
        \end{subfigure}
        \hfill
        \begin{subfigure}{.112\textwidth}
            \centering
            \includegraphics[width=\textwidth]{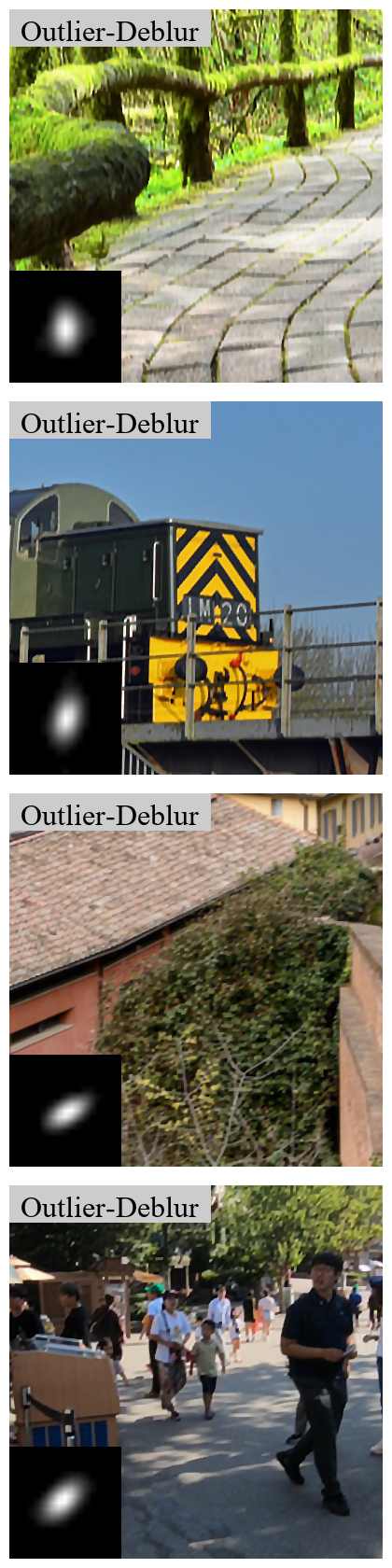}
            \caption{}
            \label{fig:deblur_a}
        \end{subfigure}
        \hfill
        \begin{subfigure}{.112\textwidth}
            \centering
            \includegraphics[width=\textwidth]{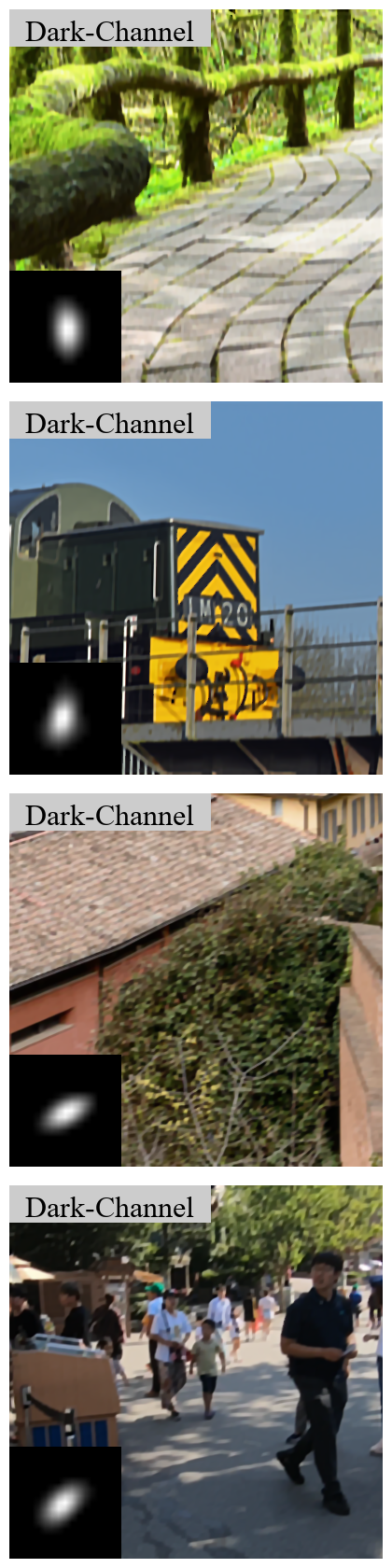}
            \caption{}
            \label{fig:deblur_a}
        \end{subfigure}
        \hfill
        \begin{subfigure}{.112\textwidth}
            \centering
            \includegraphics[width=\textwidth]{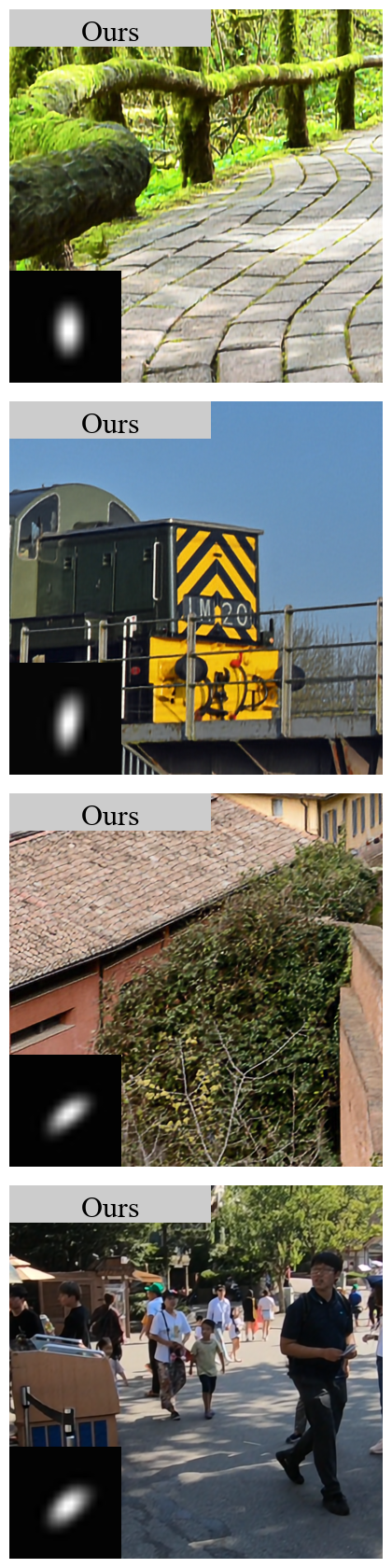}
            \caption{}
            \label{fig:deblur_a}
        \end{subfigure}
        \caption{Visualization of kernel estimation results and corresponding deblurring results. (a) Blurred image with ground-truth blur kernel. (b)--(d) Deblurred image with estimated blur kernel.}
        \label{fig:estimated_kernels}
    \end{center}
\end{figure}

To estimate the blur kernel in blind deblurring, we used our kernel estimation network shown in Figure~\ref{fig:kernel_estimator}. The blurred image we created was placed in the kernel estimation network to obtain the corresponding $17 \times 17$ blur kernel prediction result. 
Figure~\ref{fig:estimated_kernels} shows the kernel estimation results obtained using our method and several other methods. As shown, our method approximates the kernel features, such as the rotation angle, similar to other methods~\cite{pcode, darkchannel}.
During training, we first predicted the blur kernel of the blurred image and then delivered the predicted kernel to the kernel-adaptive AE block. 

For a fair comparison, we tested our model with the estimated kernel corresponding to the test blurred images, whereas ground-truth kernels were provided to other non-blind models because they required the blur kernel information. 

\subsection {Deblurring Results}

\begin{table*}[h]
\newcolumntype{?}{!{\vrule width 1.3pt}}
\setlength{\tabcolsep}{15pt}
\renewcommand{\arraystretch}{1.3}
\centering
\caption{Comparison of deblurring result on REDS, GOPRO, and Flickr2K datasets. Average PSNR and SSIM are provided with the best values marked in bold.}\label{table:deblur_result}
\begin{tabular}{lcccccc}
\thickhline
\multirow{2}{*}{Method} & \multicolumn{2}{c}{REDS} & \multicolumn{2}{c}{GOPRO} & \multicolumn{2}{c}{Flickr2K} \\ \cline{2-7}
                                & PSNR & SSIM & PSNR & SSIM & PSNR & SSIM \\ \thickhline 
Blurred Image                   & 25.9908 & 0.6420 & 28.9878 & 0.7629 & 27.6748 & 0.6750 \\
fastdeconv    & 28.4818 & 0.7644 & 33.0048 & 0.8734 & 30.2462 & 0.7838 \\
EPLL               & 30.3059 & 0.8109 & 35.0216 & 0.9058 & 31.9507 & 0.8261 \\
Outlier-Deblur               & 28.1655 & 0.7652 & 30.0099 & 0.8432 & 27.8995 & 0.7468  \\
Dark-Channel                 & 27.6806 & 0.7076 & 31.0967 & 0.8433 & 28.2186 & 0.7199    \\
DeepDeblur & 32.0374 & 0.8690 & 36.0505 & 0.9291 & 32.0912 & 0.8514 \\
SRN                  & 32.2751 & 0.8777 & 36.4768 & 0.9365 & 33.0973 & 0.8623 \\
DeblurGAN-v2 & 28.9220 & 0.8059 & 32.5397 & 0.8709 & 28.5913 & 0.7860 \\
USRNet~\cite{usrnet}                           & 30.3493 & 0.8390 & 33.5843 & 0.9255 & 32.0967 & 0.8492 \\
Ours w/o Kernel AE Block         & 32.7657 & 0.8798 & 36.6472 & 0.9351 & 32.5558 & 0.8552 \\
Ours w/o LTS Connection         & 32.7150 & 0.8786 & 36.7864 & 0.9384 & 33.2392 & 0.8630 \\
Ours                            & \textbf{33.3857} & \textbf{0.8908} & \textbf{37.2450} & \textbf{0.9437} & \textbf{33.5622} & \textbf{0.8738} \\
\thickhline
\end{tabular}
\end{table*}

We compared some methods that demonstrated excellent results with non-deep learning and deep learning methods. Fastdeconv~\cite{fastdeconv} and EPLL~\cite{epll} of the non-deep learning method and DeepDeblur~\cite{Nah_2017_CVPR}, SRN~\cite{srn}, DeblurGan-v2~\cite{deblurganv2}, and USRNet~\cite{usrnet} of the deep learning method were compared. Additionally, we compared Dark-Channel~\cite{darkchannel} and Outlier-Deblur~\cite{pcode}, which performed deblurring using the estimated kernel, similar to ours. Each method was trained and tested using the published official code. We used the PSNR and SSIM to evaluate our results.

Table~\ref{table:deblur_result} shows our experimental results. The deep learning methods indicated better results than the non-deep learning methods, and our proposed model demonstrated the best performance for all indicators. The SRN demonstrated the second best results followed by DeepDeblur with a slight difference. Because sharp images composing the GOPRO dataset were captured using a portable action camera called GOPRO, their resolutions were low. Therefore, the experiments on GOPRO yielded higher PSNR and SSIM values compared with experiments on other datasets.

As shown in Table~\ref{table:deblur_result}, the model without the kernel-adaptive AE block and long-term skip connection demonstrated a slightly lower PSNR and SSIM than our complete model. This shows that the model was better or of a similar level with a margin smaller than the SRN, which ranked top-2 in our comparative experiment. Based on this experiment, it was clear that the components of our proposed model enabled this good performance.

\begin{figure*}
    \begin{center}
        \includegraphics[width=\textwidth]{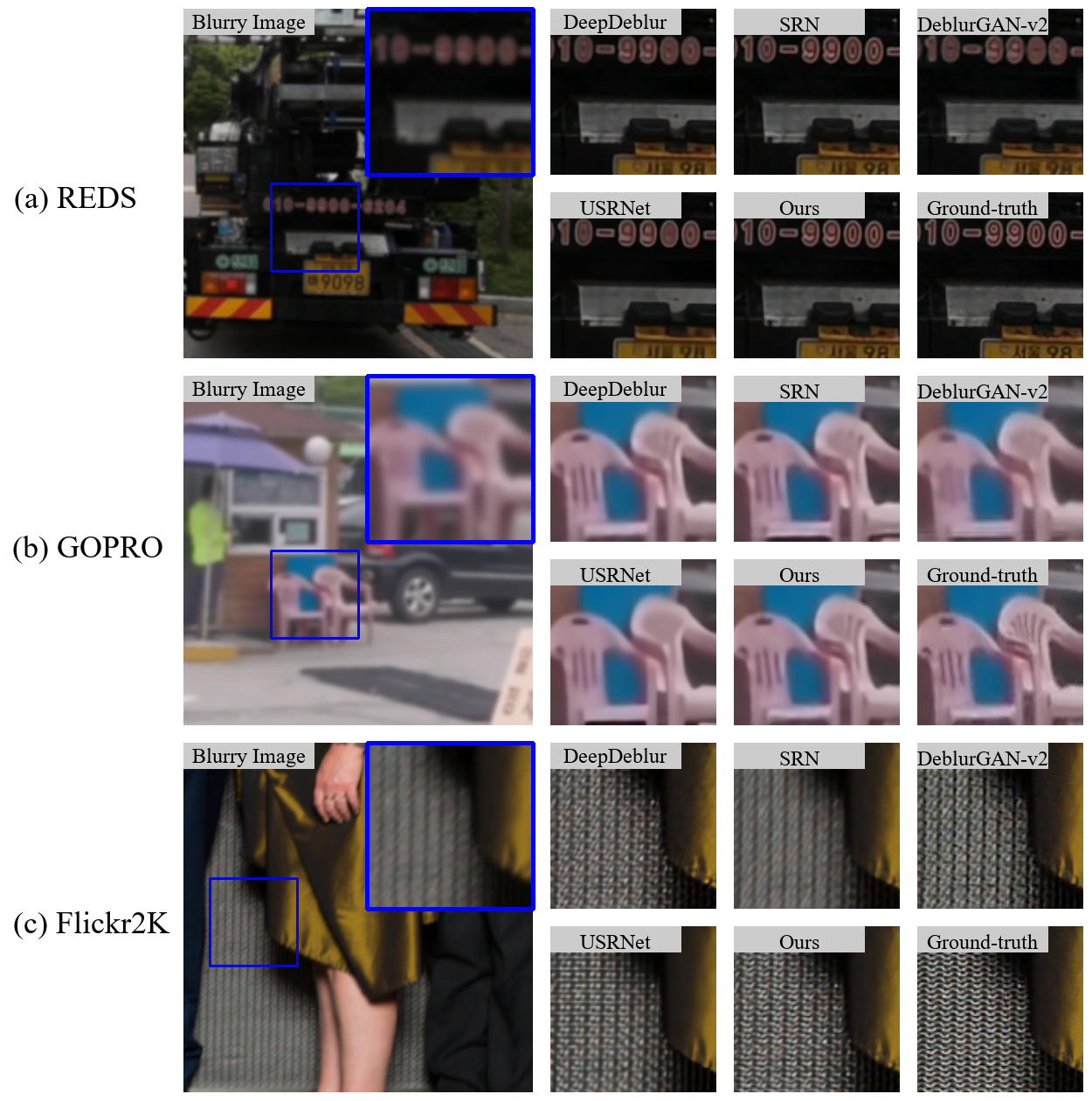}
        \caption{Deblurring results on REDS dataset}
        \caption{Visual comparison of deblurring results. In the blurred image, detailed patterns were severely damaged and difficult to recognize((a) numbers printed on the truck; (b) vertical line of the chair; (c) complex pattern on the wall). Most methods succeeded in partially restoring such patterns but more than half remained blurry. Using the blur kernel information, our method successfully restored the largest portion of damaged details and appeared the most similar to the ground-truth compared to other methods.}
        \label{fig:result_deblur}
    \end{center}
\end{figure*}

Figure~\ref{fig:result_deblur} shows detailed comparisons of qualitative results of different models on REDS, GOPRO, and Flickr2K datasets. Whereas DeepDeblur and SRN indicated better performances compared with other models, some details were still distorted and restored differently from the ground-truth. Our model, however, successfully restored such details in most cases.

\subsection{Ablation Study}
 We conducted ablation studies to evaluate the effect of our long-term skip connection and kernel-adaptive AE blocks using the estimated kernel. We first removed the kernel-adaptive AE block from our model to demonstrate its effect on our model. Subsequently, we tested our model without a long-term skip connection, the feature of which we expect to gradually utilize from coarse to fine. We used the PSNR and SSIM to evaluate our results.

As shown in Table~\ref{table:deblur_result}, the model without the kernel-adaptive AE block and long-term skip connection demonstrated a slightly lower PSNR and SSIM than our complete model. This shows that the model was better or of a similar level with a margin smaller than the SRN, which ranked top-2 in our comparative experiment. Based on this experiment, it was clear that the components of our proposed model enabled this good performance.

\subsection {Additional Experiments}

\begin{figure}[t]
    \centering
    \includegraphics[width=.474\textwidth]{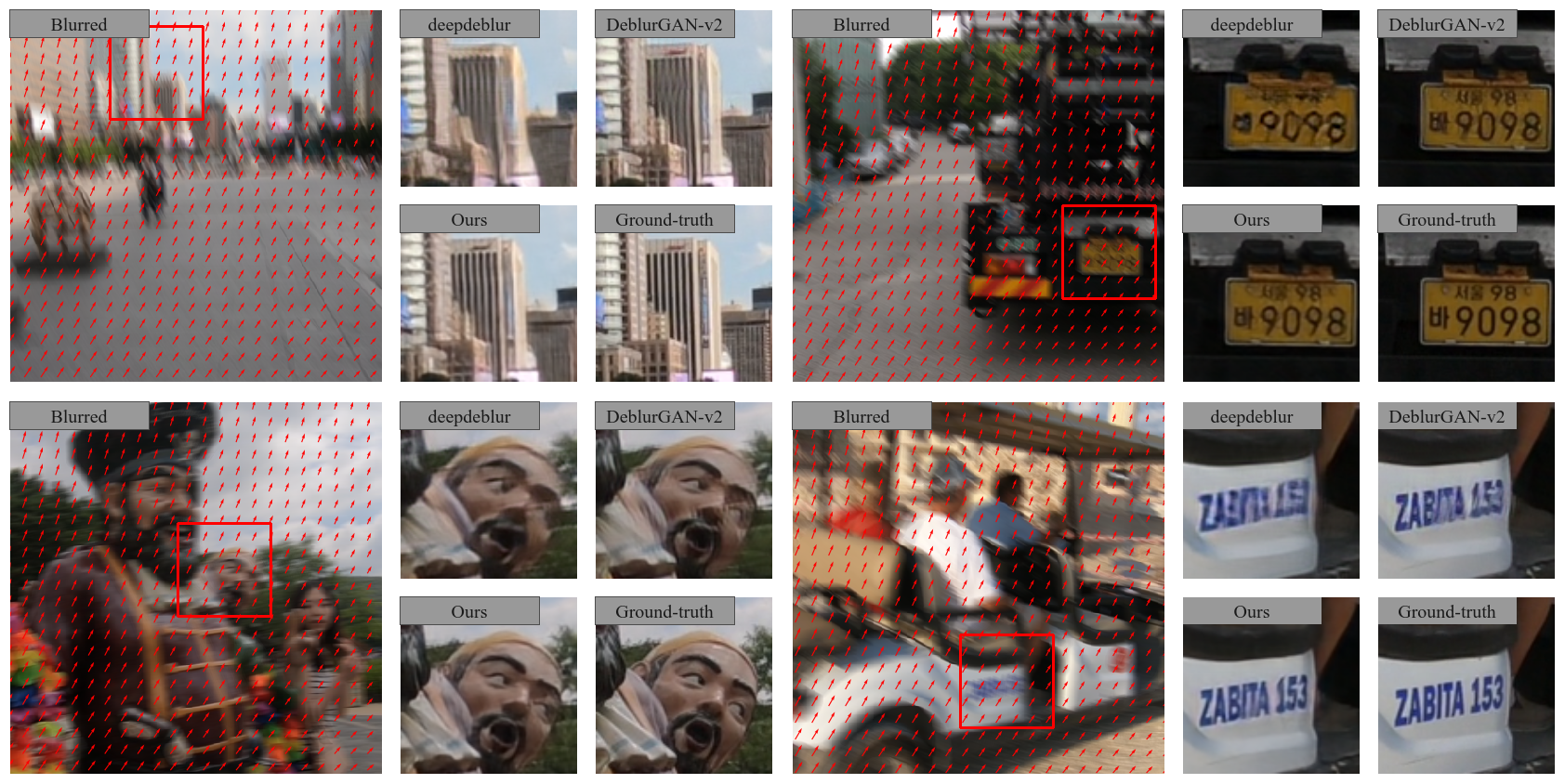}
    \caption{Visual comparison of motion deblurring results on REDS dataset}
    \label{fig:motion_blur}
\end{figure}

\setlength{\tabcolsep}{17pt}
\begin{table}[h]
\renewcommand{\arraystretch}{1.2}
\begin{center}
\caption{Comparison of motion deblurring results on REDS dataset}
\label{table:motion_blur_result}
\begin{tabular}{lll}
\thickhline
\multirow{2}{*}{Method} & \multicolumn{2}{c}{REDS}\\
\cline{2-3}
& \multicolumn{1}{c}{PSNR} & \multicolumn{1}{c}{SSIM} \\
\thickhline
Blurred Image                   & 23.5828 & 0.5117 \\
DeepDeblur  & 30.5306 & 0.8191 \\
DeblurGAN-v2 & 26.1420 & 0.6492 \\ 
Ours                            & \textbf{30.7839} & \textbf{0.8237} \\ \thickhline
\end{tabular}
\end{center}
\end{table}
\setlength{\tabcolsep}{1.4pt}

We can use our deblurring network to solve motion and Gaussian deblurring problems in a similar process: estimate the motion kernel, or the motion flow, from the blurred image and then restore the sharp image.
In particular, to solve the motion deblurring problem, we estimated the pixel-wise motion flow, such that our model can remove pixel-wise heterogeneous motion blurs.

Gong \textit{et al.}~\cite{motionflow} introduced a method for generating a dataset by translating each pixel in the $x$-, $y$-, and $z$-axes, assuming that a motion blur can be expressed as a pixel-wise motion flow. They estimated pixel-wise motion flows using a modified FCN network.
For convenience, we first assumed that the pixel-wise motion flow was that assumed by Gong, \textit{et al.} Subsequently, we trained and evaluated our BIKAnet using that pixel-wise motion flow in a non-blind manner. For the dataset, we generated blurred images from sharp images in the REDS dataset using the blurring method proposed by Gong \textit{et al.}

To create motion flows in our model and for the motion deblurring task, we modified the structure of the encoder and decoder modules in the kernel-adaptive AE blocks from the BIKAnet.
Because motion flows are estimated pixel-wise, unlike the Gaussian blur kernel, we concatenated the motion flow and the feature instead of AdaIN and then placed them into the kernel-adaptive AE block as a condition.

Figure~\ref{fig:motion_blur} and Table~\ref{table:motion_blur_result} show a comparison of the results obtained using our model and other state-of-the-art methods. In terms of the PSNR and SSIM, Table~\ref{table:motion_blur_result} shows that our model performed better than the other methods. Based on the results of the reconstructed image in Figure~\ref{fig:motion_blur}, our model and DeepDeblur performed well together; however, the result of DeepDeblur showed some artifacts and insufficient details. Our model not only performed well on the deblurring task, but also demonstrated good results in the motion deblurring domain. We expect our model to perform better if we further change the design of the encoder that projects the features of motion flow to render our model more optimized for the motion deblurring task. However, this will only be attempted in future research. 

\section{Conclusion}

In this paper, we presented a novel deblurring method. Using our proposed kernel-adaptive AE block and long-term skip connection, we improved the understanding of the image deblurring process and the method to use blur kernel estimation efficiently. Our model not only demonstrated excellent results in the deblurring task, but it also indicated the potential to be applied to other image enhancement tasks such as motion deblurring.


\bibliography{main}
\end{document}